# Missing Data Reconstruction in Remote Sensing image with a Unified Spatial-Temporal-Spectral Deep Convolutional Neural Network

Qiang Zhang, *Student Member, IEEE*, Qiangqiang Yuan, *Member, IEEE*, Chao Zeng, Xinghua Li, Yancong Wei

*Abstract*—Because of the internal malfunction of satellite sensors and poor atmospheric conditions such as thick cloud, the acquired remote sensing data often suffer from missing information, i.e., the data usability is greatly reduced. In this paper, a novel method of missing information reconstruction in remote sensing images is proposed. The unified spatial-temporal-spectral framework based on a deep convolutional neural network (STS-CNN) employs a unified deep convolutional neural network combined with spatial-temporal-spectral supplementary information. In addition, to address the fact that most methods can only deal with a single missing information reconstruction task, the proposed approach can solve three typical missing information reconstruction tasks: 1) dead lines in Aqua MODIS band 6; 2) the Landsat ETM+ Scan Line Corrector (SLC)-off problem; and 3) thick cloud removal. It should be noted that the proposed model can use multi-source data (spatial, spectral, and temporal) as the input of the unified framework. The results of both simulated and real-data experiments demonstrate that the proposed model exhibits high effectiveness in the three missing information reconstruction tasks listed above.

## I. INTRODUCTION

The Earth observation technology of remote sensing is one of the most important ways to obtain geometric attributes and physical properties of the Earth's surface. However, because of the satellite sensor working conditions and the atmospheric environment, remote sensing images often suffer from missing information problems, such as dead pixels and thick cloud cover [1], as shown in Fig. 1.

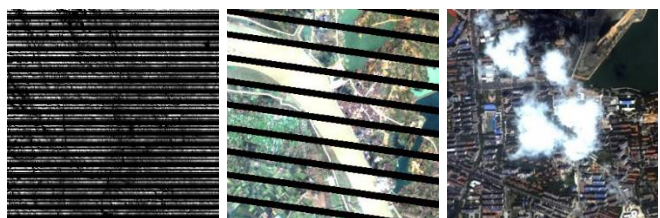

(a) Dead lines in Aqua MODIS band 6   (b) Landsat ETM+ SLC-off   (c) QuickBird image with thick cloud cover

Fig. 1. The traditional missing information problems of remote sensing data.

To date, a variety of missing information reconstruction methods for remote sensing imagery have been proposed. According to the information source, most of the reconstruction methods can be classified into four main categories [1]: **1) spatial-based methods; 2) spectral-based methods; 3) temporal-based methods;** and **4) spatial-temporal-spectral-based methods.** Details of these methods are provided in the discussion in Section II. Although these different approaches can acquire satisfactory recovery results, most of them are employed independently, and they can only be applied to a single specific reconstruction task in limited conditions [4]. Therefore, it is worth proposing a unified missing data reconstruction framework which can jointly take advantage of auxiliary complementary data from the spatial, spectral, and temporal domains, for different missing information tasks, such as the dead lines of the Aqua Moderate Resolution Imaging Spectroradiometer (MODIS) band 6, the Landsat Enhanced Thematic Mapper Plus (ETM+) Scan Line Corrector (SLC)-off problem, and thick cloud removal. Furthermore, most of the existing methods are based on linear models, and thus have difficulty dealing with complex scenarios and reconstructing large missing areas. Therefore, innovative ideas need to be considered to break through the constraints and shortcomings of the traditional methods.

Recently, benefiting from the powerful non-linear expression ability of deep learning theory [5], convolutional neural networks (CNNs) [6] have been successfully applied to many low-level vision tasks for remote sensing imagery, such as optical remote sensing image super-resolution [7], hyperspectral image denoising [8], pansharpening [9], and so on [10]–[13]. Therefore, in this paper, from the perspective of deep learning theory and spatial-temporal-spectral fusion [14], we propose a unified spatial-temporal-spectral framework based on a deep convolutional neural network (STS-CNN) for the reconstruction of remote sensing imagery contaminated with dead pixels and thick cloud. It should be noted that the proposed method can use multi-source data (spatial, spectral, and temporal) as the input of the unified framework. The results of both simulated and real-data experiments suggest that the proposed STS-CNN model exhibits a high effectiveness in the three reconstruction tasks listed above. The main contributions can be summarized as follows.

1) A novel deep learning based-method is presented for reconstructing missing information in remote sensing imagery.

The proposed method learns a non-linear end-to-end mapping between the missing data and intact data with auxiliary data through a deep convolution neural network. In the proposed model, we employed residual output instead of straightforward output to learn the relations between different auxiliary data. The learning procedure with residual unit is much sparse, and easier to approximate to the original data through the deeper and intrinsic feature extraction and expression.

2) We proposed a unified multi-source data framework combined with spatial-temporal-spectral supplementary information to boost the recovering accuracy and consistency. It should be noted that, the proposed model can use multiple data (spatial, spectral, and temporal) as the input of the unified framework with the deep convolution neural network for different reconstructing tasks.

3) To address the deficiency that most methods can only deal with a single missing information reconstruction task, the proposed approach shows the universality of various missing information reconstruction tasks like: 1) dead lines in Aqua MODIS band 6; 2) the Landsat ETM+ SLC-off problem; and 3) thick cloud removal. The simulated and real experiments manifest that the proposed STS-CNN outperforms many current mainstream methods in both evaluation indexes and visual reconstructing perception.

The remainder of this paper is organized as follows. The related works about the pre-existing methods of missing information reconstruction in remote sensing imagery are introduced in Section II. The network architecture and specific details of the proposed STS-CNN model are described in Section III. The results of the missing data reconstruction in both simulated and real-data experiments are presented in Section IV. Finally, our conclusions and expectations are summarized in Section V.

## II. RELATED WORK

### A. Spatial-Based Methods

The spatial-based methods, which are also called "image inpainting" methods, are the most basic methods in image reconstruction in the field of computer vision. These methods usually assume that undamaged regions have the same or related statistical features or texture information as the missing regions. In addition, the spatial relationship between the global and local areas may also be considered in the reconstruction procedure. The spatial-based methods include interpolation methods [15]–[16], exemplar-based methods [17]–[19], partial differential equation (PDE)-based methods [20]–[21], variational methods [22]–[23], and learning-based method [24]–[25]. For example, the interpolation methods seek the weighted average of pixels of the neighborhood area around the missing region, which is the most commonly used method. The advantage of the interpolation methods is that they are easy and efficient, but they cannot be applied to the reconstruction of large missing areas or areas with complex texture. Therefore, to solve this problem, some new strategies have been presented, such as PDE-based methods and exemplar-based methods. Nevertheless, the application scenarios of these methods are restricted by the specific texture structure and the size of the missing areas. Recently, with the development of deep learning, Pathak *et al.* [24] used an encoder-decoder CNN and adversarial loss to recover missing regions; and Yang *et al.* [25] further used Markov random fields (MRFs) to constrain the texture feature and improve the spatial resolution. However, these methods still cannot solve the problem of reconstructing large areas with a high level of precision.

In general, the spatial-based methods are qualified for reconstructing small missing areas or regions with regular texture. However, the reconstruction precision cannot be guaranteed, especially for large or complex texture areas.

### B. Spectral-Based Methods

To overcome the bottleneck of the spatial-based methods, adding spectral information to the reconstruction of missing data provides another solution. For multispectral or hyperspectral imagery, there is usually high spatial correlation between the different spectral data, which provides the possibility to reconstruct the missing data based on the spectral information.

For examples, since Terra MODIS bands 6 and 7 are closely correlated, Wang *et al.* [26] employed a polynomial linear fitting (LF) method between the data of Aqua MODIS bands 6 and 7, whose missing data could be obtained by this linear fit formula. Based on this idea, Rakwatin *et al.* [27] presented an algorithm combining histogram matching with local least squares fitting (HMLLSF) to reconstruct the missing data of Aqua MODIS band 6. Subsequently, Shen *et al.* [28] further developed a within-class local fitting (WCLF) algorithm, which additionally considers that the band relationship is relevant to the scene category. Furthermore, Li *et al.* [29] employed a robust M-estimator multi-regression (RMEMR) method based on the spectral relations between working detectors in Aqua MODIS band 6 and all the other spectra to recover the missing information of band 6.

In conclusion, the spectral-based methods can recover the missing spectral data with a high level of accuracy through employing the high correlation between the different spectral data. However, these methods cannot deal with thick cloud cover, because this leads to the absence of all the spectral bands, to different degrees.

### C. Temporal-Based Methods

Temporal information can also be utilized to recover the missing data, on account of the fact that satellites can obtain remote sensing data in the same region at different times. Therefore, the temporal-based methods are reliant on the fact that time-series data are strictly chronological and display regular fluctuations. For instance, Scaramuzza *et al.* [30] presented a local linear histogram matching (LLHM) method, which is simple to realize and can work well in most areas if the input data and auxiliary data are of high quality. However, it often obtains poor results, especially for heterogeneous landscapes, where the feature size is smaller than the local moving window size. Chen *et al.* [31] put forward a simple approach known as neighborhood similar pixel interpolation (NSPI), through combining local area replacement and interpolation, which can even fill the SLC-off gaps in non-uniform regions. Zeng *et al.* [32] proposed a weighted linear regression (WLR) method for reconstructing missing data, using multi-temporal images as referable information and then building a regression model between the corresponding missing pixels. Furthermore, Li *et al.* [33] established a relationship map between the original and temporal data, with multi-temporal dictionary learning based on sparse representation. Zhang *et al.* [34] presented a functional concurrent linear model (FCLM) to address missing data problems in series of temporal images. Chen *et al.* [35] developed a novel spatially and temporally weighted regression (STWR) model for cloud removal to produce continuous cloud-free Landsat images. Besides, Gao *et al.* [36] proposed tempo-spectral angle mapping (TSAM) method for SLC-off to measure tempo-spectral similarity between pixels described in spectral dimension and temporal dimension.

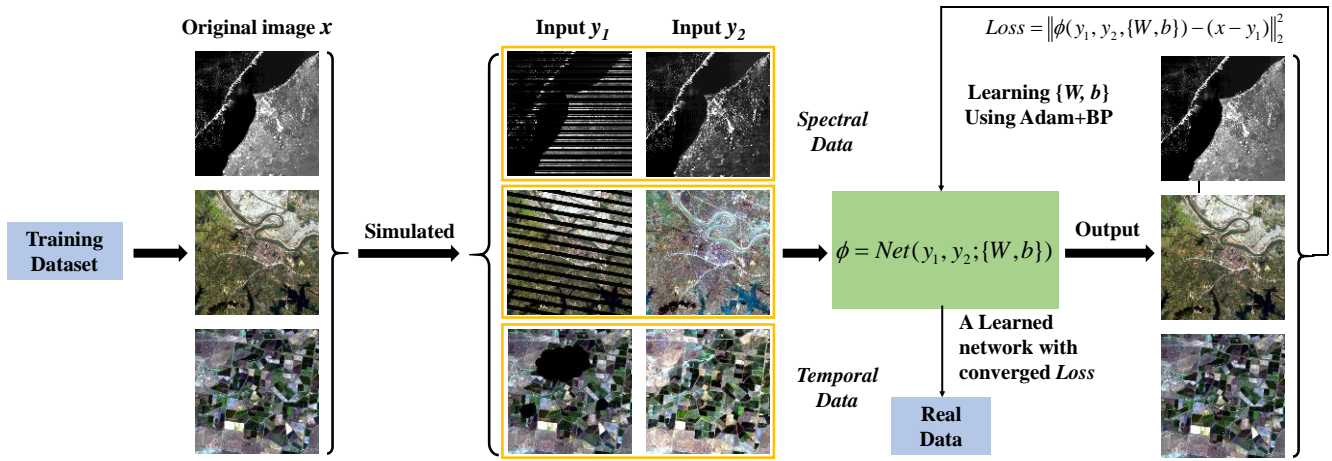

Fig. 2. Flowchart of the STS-CNN framework for the missing information reconstruction of remote sensing imagery.

In summary, for the temporal-based methods, although they can work well for a variety of situations such as thick cloud and ETM+ SLC-off, the temporal differences are major obstacles to the reconstruction process, and registration errors between multi-temporal images also have a negative impact on the precision of the corresponding recovered regions.

### D. *Spatial-Temporal-Spectral-Based Methods*

Despite the fact that many types of methods for the reconstruction of missing information in remote sensing imagery have been proposed, most of them have been developed independently for a single recovery task. However, a few researchers have attempted to explore a unified framework to deal with the different missing information tasks with spatial, temporal, and spectral complementary information. For example, Ng *et al.* [4] proposed a single-weighted low-rank tensor (AWTC) method for the recovery of remote sensing images with missing data, which collectively makes use of the spatial, spectral, and temporal information in each dimension, to build an adaptive weighted tensor low-rank regularization model for recovering the missing data. Besides, Li *et al.* [37] also presented a spatio-spectral-temporal approach for the missing information reconstruction of remote sensing images based on group sparse representation, which utilizes the spatial correlations from local regions to non-local regions, by extending single patch based sparse representation to multiple patch based sparse representation.

Beyond that, the highly non-linear spatial relationship between multi-source remote sensing images indicates that higher-level expression and better feature representation are essential for the reconstruction of missing information. However, most of the methods based on linear models cannot deal well with complex non-linear degradation models, such as image inpainting, super-resolution, denoising etc. Therefore, the powerful non-linear expression ability of deep learning (e.g., CNNs) can be introduced for recovering degraded images.

To date, to the best of our knowledge, no studies investigating CNNs for the reconstruction of missing information in remote sensing imagery have made full use of the feature mining and non-linear expression ability. Therefore, we propose a novel method from the perspective of a deep CNN combined with joint spatial-temporal-spectral information, which can solve all three-typical missing information reconstruction tasks: 1) the dead lines of Aqua MODIS band 6; 2) the Landsat SLC-off problem; and 3) thick cloud cover. The overall framework and details of the proposed method are provided in Section III.

### III. PROPOSED RECONSTRUCTION FRAMEWORK

#### A. *Fundamental Theory of CNNs*

With the recent advances made by deep learning for computer vision and image processing applications, CNNs have gradually become an efficient tool which has been successfully applied to many computer vision tasks, such as image classification, segmentation, object recognition, and so on [5]. CNNs can extract the internal and underlying features of images and avoid complex *a priori* constraints. CNNs are organized in a feature map $O_j^{(l)}$ ( $j=1,2,\ldots M^{(l)}$ ), within which each unit is connected to local patches of the previous layer $O_j^{(l-1)}$ ( $j=1,2,\ldots M^{(l-1)}$ ) through a set of weight parameters $W_j^{(l)}$ and bias parameters $b_j^{(l)}$. The output feature map is:

$$L_j^{(l)}(m,n) = F(O_j^{(l)}(m,n)) \quad (1)$$

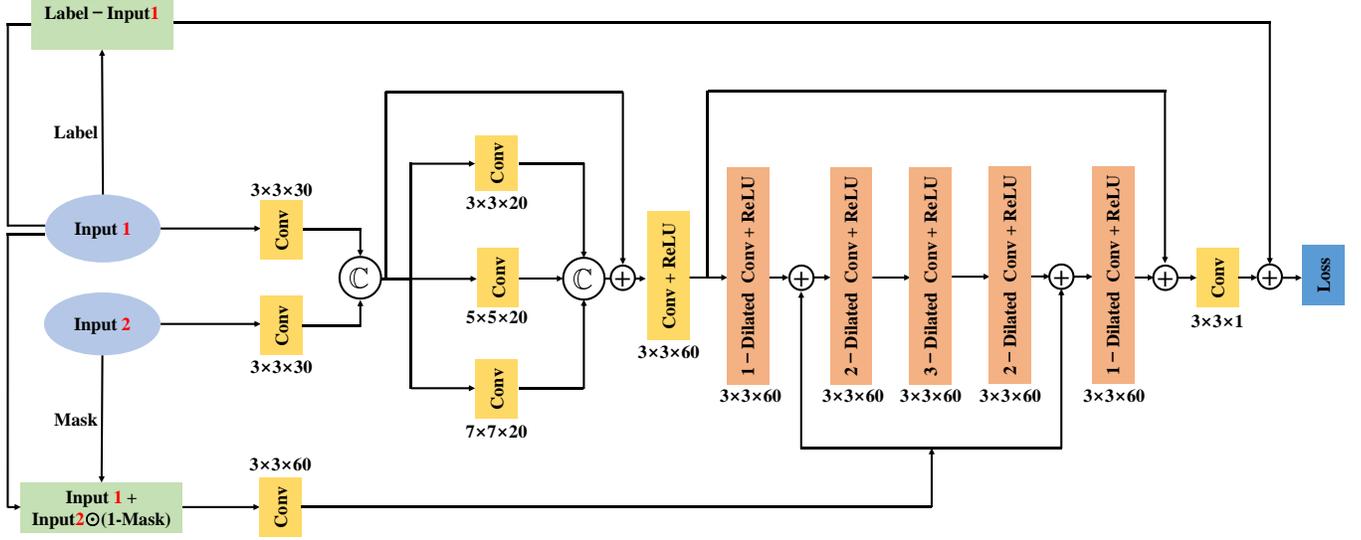

Fig. 3. The architecture of the proposed STS-CNN framework.

And

$$O_j^{(l)}(m,n) = \sum_{i=1}^{M^{(l)}} \sum_{u,v=0}^{S-1} W_{ji}^{(l)}(u,v) \cdot L_i^{(l-1)}(m-u,n-v) + b_j^{(l)} \quad (2)$$

where $F(\cdot)$ is the non-linear activation function, and $O_j^{(l)}(m,n)$ represents the convolutional weighted sum of the previous layer's results to the $j$ th output feature map at pixel $(m,n)$. Furthermore, the special parameters in the convolutional layer include the number of output feature maps $j$ and the filter kernel size $S \times S$. In particular, the network parameters $W$ and $b$ need to be regenerated through back-propagation and the chain rule of derivation [6].

To ensure that the output of the CNN is a non-linear combination of the input, due to the fact that the relationship between the input data and the output label is usually a highly non-linear mapping, a non-linear function is introduced as an excitation function. For example, the rectified linear unit (ReLU) is defined as:

$$F(O_j^{(l)}) = \max(0, O_j^{(l)}) \quad (3)$$

After finishing each process of the forward propagation, the back-propagation algorithm is used to update the network parameters, to better learn the relationships between the labeled data and reconstructed data. The partial derivative of the loss function with respect to convolutional kernels $W_{ji}^{(l)}$ and bias $b_j^{(l)}$ of the $l$ th convolutional layer is respectively calculated as follows:

$$\frac{\partial L}{\partial W_{ji}^{(l)}} = \sum_{m,n} \delta_j^{(l)}(m,n) \cdot L_j^{(L-1)}(m-u, y-v) \quad (4)$$

$$\frac{\partial L}{\partial b_j^{(l)}} = \sum_{m,n} \delta_j^{(l)}(m,n) \quad (5)$$

where the error map $\delta_j^{(l)}$ is defined as:

$$\delta_j^{(l)} = \sum_j \sum_{u,v=0}^{S-1} W_{ji}^{(l+1)}(u,v) \cdot \delta_j^{(l+1)}(m+u, n+v) \quad (6)$$

The iterative training rule for updating the network parameters $W_{ji}^{(l)}$ and $b_j^{(l)}$ through the gradient descent strategy is as follows:

$$W_{ji}^{(l)} = W_{ji}^{(l)} + \alpha \cdot \frac{\partial L}{\partial W_{ji}^{(l)}} \quad (7)$$

$$b_j^{(l)} = b_j^{(l)} + \alpha \cdot \frac{\partial L}{\partial b_j^{(l)}} \quad (8)$$

where $\alpha$ is a hyperparameter for the whole network, which is also named the "learning rate" in the deep learning framework.

### B. The Whole Framework Description

Aiming at the fact that most methods can only deal with a single type of missing information reconstruction, the proposed framework can simultaneously recover dead pixels and remove thick cloud in remote sensing images. The STS-CNN framework is depicted in Fig. 2.

To learn the complicated non-linear relationship between input $y_1$ (spatial data with missing regions) and input $y_2$ (auxiliary spectral or temporal data), the proposed STS-CNN model is employed with converged loss between the original image $x$ and the input $y_1$. The full details of this network are provided in Section III-C.

## C. The Proposed STS-CNN Reconstruction Framework

Inspired by the basic idea of the image fusion strategy to boost the spatial resolution, the proposed STS-CNN framework introduces several structures to enhance the manifestation of the proposed network. The overall architecture of the STS-CNN framework is displayed in Fig. 3. The label data in the proposed model is the original image without missing data as shown in the flowchart in Fig. 2. Detailed descriptions of each component of STS-CNN are provided in the following.

### 1) Fusion of Multi-Source Data

As mentioned in Section II, complementary information, such as spectral or temporal data, can greatly help to improve the precision of the reconstruction as such data usually have a high correlation with the missing regions in the surface properties and textural features. Therefore, in the proposed STS-CNN framework, we input two types of data into the network, one of which is the spatial data with missing areas (input $y_1$ in Fig. 4), and the other is the complementary information, such as spectral or temporal data (input $y_2$ in Fig. 4).

For the dead lines in Aqua MODIS band 6, input $y_1$ is the spectral data with missing information, and input $y_2$ is the other intact spectral data as auxiliary information. For the ETM+ SLC-off problem, input $y_1$ is the temporal image with missing information, and input $y_2$ is another temporal image. For removing thick cloud in remote sensing imagery, input $y_1$ is a temporal image with regions covered by thick cloud, and input $y_2$ is another temporal image without cloud.

The two inputs respectively go through one layer of convolution operation with a 3×3 kennel size, and generate an output of 30 feature maps, respectively. The two outputs of feature maps are then concatenated to the size of 3×3×60, as shown in Fig. 4.

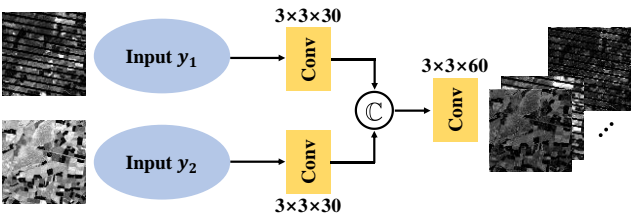

Fig. 4. Fusion of multi-source data with convolutional layers and a concatenation layer.

### 2) Multi-Scale Convolutional Feature Extraction Unit

In the procedure for reconstructing the missing information in remote sensing imagery, the procedure may rely on contextual information in different scales, due to the fact that ground objects usually have multiplicative sizes in different non-local regions. Therefore, the proposed model introduces a multi-scale convolutional unit to extract more features for the multi-context information. As shown in Fig. 5(a), the multi-scale convolutional unit contains three convolution operations of 3×3, 5×5, and 7×7 kernel sizes, respectively. All three convolutions are simultaneously conducted on the feature maps of the input data, and produce feature maps of 20 channels, as shown in Fig. 5(b). The three feature maps are then concatenated into a single 60-channel feature map, such that the features extracting the contextual information with different scales are fused together for posterior processing.

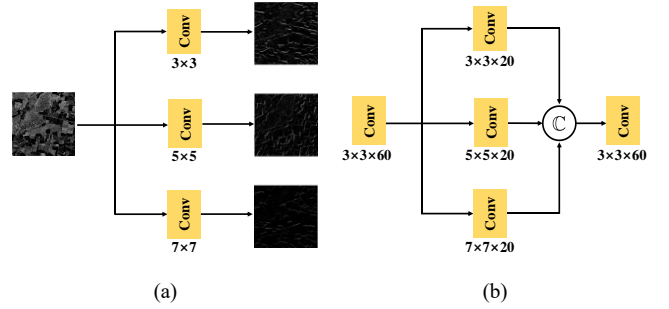

Fig. 5. Multi-scale convolution feature extraction block. (a) Example of multi-scale convolution operations of 3×3, 5×5, and 7×7 kernel sizes. (b) Integral structure of the multi-scale convolution feature extraction block in STS-CNN.

### 3) Dilated Convolution

In image inverse problems such as image inpainting [38], denoising [39]–[40], and deblurring [41], contextual information can effectively promote the restoration of degraded images. Similarly, in deep CNNs, it enhances the contextual information through enlarging the receptive field during the convolution operations. In general, there are two strategies to reach this target: 1) increasing the layers of the network; and 2) enlarging the size of the convolution kernel filter. However, on the one hand, as the network depth increases, the accuracy becomes "saturated" and then rapidly degrades due to the back-propagation. On the other hand, enlarging the size of the kernel filter can also introduce convolution parameters, which greatly increases the calculative burden and training times.

To solve this issue effectively, dilated convolutions are employed in the STS-CNN model, which can both enlarge the receptive field and maintain the size of the convolution kernel filter. Differing from common convolution, the dilated convolution operator can employ the same filter at different ranges using different dilation factors. Setting the kernel size as 3×3 as an example, Fig. 6 illustrates the dilated convolution receptive field size by the green color. The common convolution receptive field has a linear correlation with the layer depth, in that the receptive field size $F_{depth-i} = (2i+1) \times (2i+1)$, while the dilated convolution receptive field has an exponential correlation with the layer depth, which is equal to $F_{depth-i} = (2^{i+1}-1) \times (2^{i+1}-1)$. For the reconstruction model, the dilation factors of the 3×3 dilated convolutions from layer 2 to layer 6 are respectively set to 1, 2, 3, 2, and 1, as shown in Fig. 7.

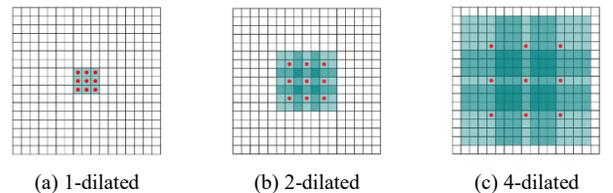

(a) 1-dilated　　(b) 2-dilated　　(c) 4-dilated

Fig. 6. The receptive field size (1, 2 and 4) with dilated convolution.

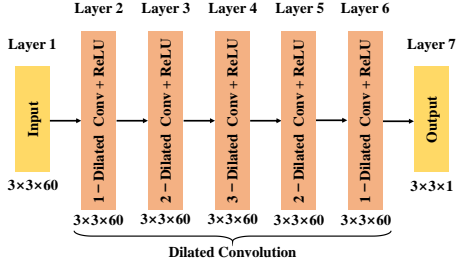

Fig. 7. The dilated convolution in the proposed network.

### 4) Boosting of the Spatial-Temporal-Spectral Information

To maintain and boost the transmitting of the spatial and spectral/temporal information in the proposed method, a unique structure was specially designed, as shown in Fig. 8.

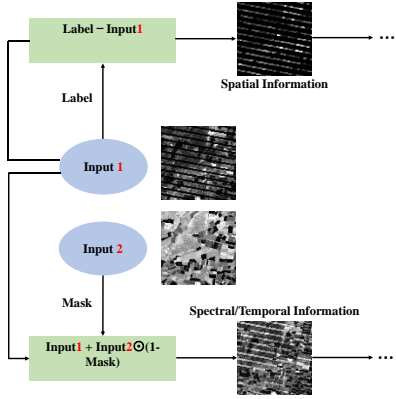

Fig. 8. Boosting of the spatial and spectral/temporal information.

To preserve the spatial information, the residual image between the label and *input 1* is transferred to the last layer before the loss function, which is also equivalent to the constructed part of the missing regions. As our input data and output results are largely the same in intact regions, we define a residual mapping:

$$r_i = y_i^1 - x_i \quad (9)$$

where $y_i^1$ (the *input 1* in Fig. 3) is the image with missing data, and $x_i$ is the original undamaged image. Compared with traditional data mapping, this residual mapping can acquire a more effective learning status and rapidly reduce the training loss after passing through a multi-layer network. In particular, $r_i$ is also just equivalent to the missing regions, outside which most pixel values in the residual image are close to zero, and the spatial distribution of the residual feature maps should be very sparse, which can transfer the gradient descent process to a much smoother hyper-surface of loss to the filtering parameters. Thus, searching for an allocation which is on the verge of the optimal for the network's parameters becomes much quicker and easier, allowing us to add more layers to the network and improve its performance.

Specifically for the proposed model, given a collection of $N$ training image pairs $\{x_i, y_i^1, y_i^2\}_N$, $y_i^2$ (the *input 2* in Fig. 3) is the spectral or temporal auxiliary image, and $\Theta$ is the network parameters. The mean squared error (MSE) as the loss function in the proposed model is defined as:

$$loss(\Theta) = \frac{1}{2N}\sum_{i=1}^{N}\left\|\phi(y_i^1, y_i^2, \Theta) - r_i\right\|_2^2 \quad (10)$$

Furthermore, to ensure the spectral/temporal information and reduce the spectral distortion, *input 1* with filled missing gaps by *input 2* and *mask* is transferred to the subsequent layer in the network, which can also enhance the feature of auxiliary spectral/temporal information as the data transferring with multi-layers, as shown in Fig. 8.

### 5) Skip Connection

Although the increase of the network layer depth can help to obtain more data feature expressions, it often results in the gradient vanishing or exploding problem, which causes the training of the model to be much harder. To solve or reduce this problem, a new structure called the skip connection is employed for the deep CNN. The skip connection can pass the previous layer's feature information to its posterior layer, maintaining the image details and avoiding or reducing the vanishing gradient problem. In the proposed STS-CNN model, three skip connections are employed in the multi-scale convolution block (as shown in Fig. 9(a)), where the input and output of the dilated convolution (upper solid line in Fig. 9 (b)) and the foregoing feature maps of the spectral/temporal information connect with the feature maps after the first and fourth dilated convolutions (lower solid line in Fig. 9 (b)).

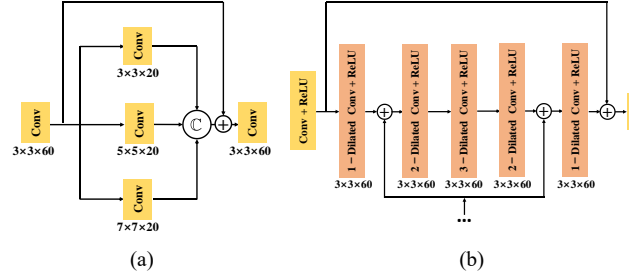

Fig. 9. Skip connections in the proposed model. (a) Skip connection in the multi-scale convolution unit. (b) Skip connection in the dilated convolution.

## IV. EXPERIMENTAL RESULTS AND DISCUSSION

### A. Experimental Settings

In this study, we used the single framework to solve three tasks mentioned above. For different reconstructing tasks, the corresponding training data are employed independently to train the specific models, respectively. Details of our experimental settings are given as below.

**1) Training and Test Data:** For the dead lines of Aqua MODIS Band 6, we selected original Terra MODIS imagery from [42] as our training dataset, since it has a high degree of similarity. For the training of the network, we chose and cropped 600 images of size 400×400×7 and set each patch size as 40×40 and stride = 40. To test the performance of the proposed model, another single example of the Terra MODIS image was set up as a simulated image. In addition, for the real dead lines of Aqua MODIS band 6, an Aqua MODIS L1B 500-m resolution image of size 400×400×7 was used in the real-data experiments.

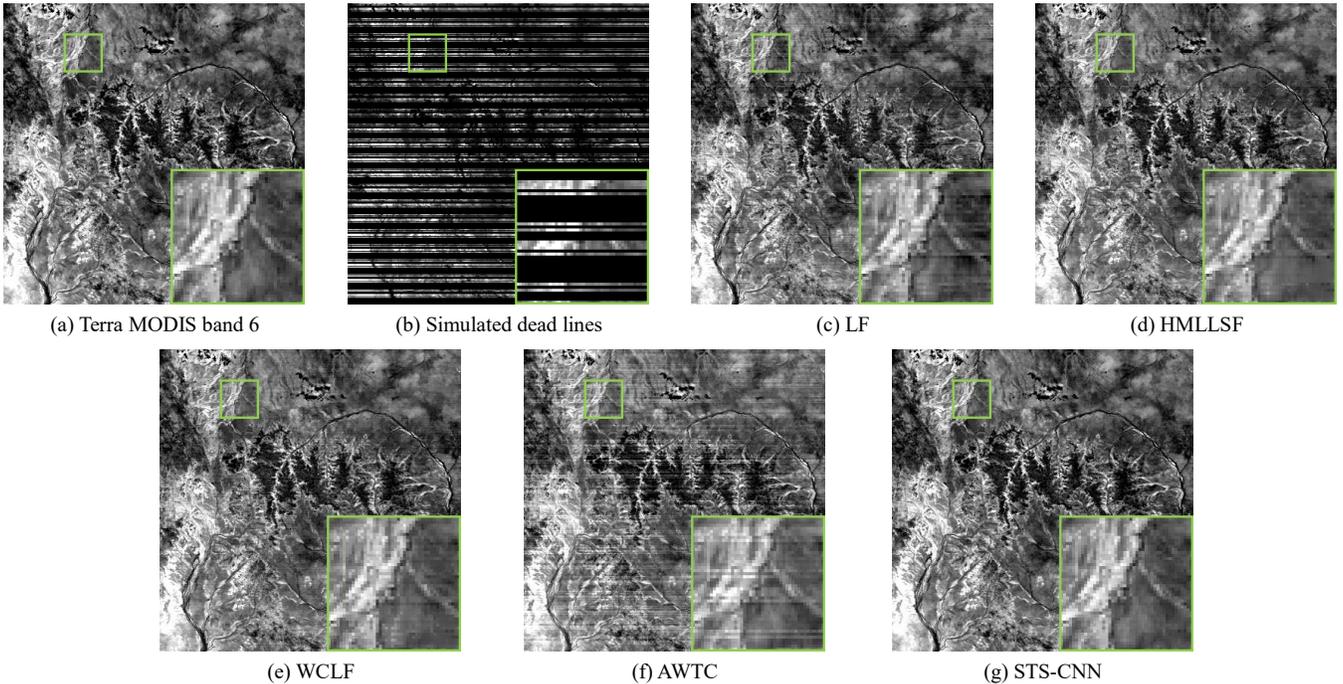

Fig. 10. Simulated recovery results of Terra MODIS band 6.

For the ETM+ SLC-off problem and the removal of thick cloud, we used 16 different temporal Landsat Thematic Mapper (TM) images from 2001.10.7 to 2002.5.4 (size of 1720×2040×6, 30-m spatial resolution) and arranged them in sets of temporal pairs. These pairs of temporal data were then cropped in each patch size as 100×100 and stride = 100 as the training datasets. For the SLC-off problem and the removal of thick cloud, another two single examples of two temporal Landsat images (400×400×6) were set up as simulated images with a missing information mask. Two actual ETM+ SLC-off temporal images, and two temporal images with/without cloud were also tested for the real-data experiments. For all the simulated experiments through different algorithms, we repeated the reconstructing procedures with 10 times. Mean and standard deviation values of the evaluation indexes are listed in Tables I-IV, respectively.

**2) Parameter Setting and Network Training:** The proposed model was trained using the stochastic gradient descent (SGD) [43] algorithm as the gradient descent optimization method, where the learning rate $\alpha$ was initialized to 0.01 for the whole network. For the different reconstruction tasks, the training processes were all set to 100 epochs. After every 20 epochs, the learning rate $\alpha$ was multiplied by a declining factor $gamma = 0.1$. In addition, the proposed network employed the *Caffe* [44] framework to train in the Windows 7 environment, with 16-GB RAM, an Intel Xeon E5-2609 v3 @1.90 GHz CPU, and an NVIDIA TITAN X (Pascal) GPU. The testing codes of STS-CNN can be downloaded at https://github.com/WHUQZhang/STS-CNN.

**3) Compared Algorithms and Evaluation Indexes:** For the dead lines in Aqua MODIS band 6, the proposed method was compared with four algorithms: polynomial LF [26], HMLLSF [27], WCLF [28] and AWTC [4]. The peak signal-to-noise ratio (PSNR), the structural similarity (SSIM) index [45], and the correlation coefficients (CCs) were employed as the evaluation indexes in the simulated experiments. For the ETM+ SLC-off problem, the proposed method was compared with five algorithms: LLHM [30], NSPI [31] WLR [32], TSAM [36], and AWTC [4]. For the removal of thick cloud, the proposed method was compared with five algorithms: LLHM [30], mNSPI [46], WLR [32], STWR [35], and AWTC [4]. The mean PSNR and SSIM (mPSNR and mSSIM) values of all the spectral bands, CCs, and spectral angle mapper (SAM) [14] were employed as the evaluation indexes in the simulated experiments.

### B. Simulated Experiments

**1) Simulated Dead Lines in Terra MODIS Band 6**

The MODIS sensors on both the Aqua and Terra satellites have similar design patterns, which makes it possible to consider the reconstruction result of the simulated dead lines in Terra MODIS as the approximate evaluation approach [1]. In Fig. 10, we present the simulated recovery results of Terra MODIS band 6 through the five methods: LF [26], HMLLSF [27], WCLF [28], AWTC [4], and the proposed STS-CNN. In addition, the quantitative evaluations with PSNR, SSIM, and CC are shown in Table I.

TABLE I. QUANTITATIVE EVALUATION RESULTS OF THE SIMULATED DEAD LINES IN TERRA MODIS BAND 6.

|  | PSNR/dB (↑) | SSIM (↑) | CC (↑) |
| --- | --- | --- | --- |
| LF | 40.236 ± 0.000 | 0.971 ± 0.000 | 0.967 ± 0.000 |
| HMLLSF | 42.787 ± 0.014 | 0.984 ± 0.001 | 0.982 ± 0.001 |
| WCLF | 43.498 ± 0.025 | 0.986 ± 0.002 | 0.985 ± 0.002 |
| AWTC | 39.984 ± 0.012 | 0.975 ± 0.001 | 0.962 ± 0.001 |
| STS-CNN | **44.569 ± 0.018** | **0.990 ± 0.001** | **0.989 ± 0.001** |

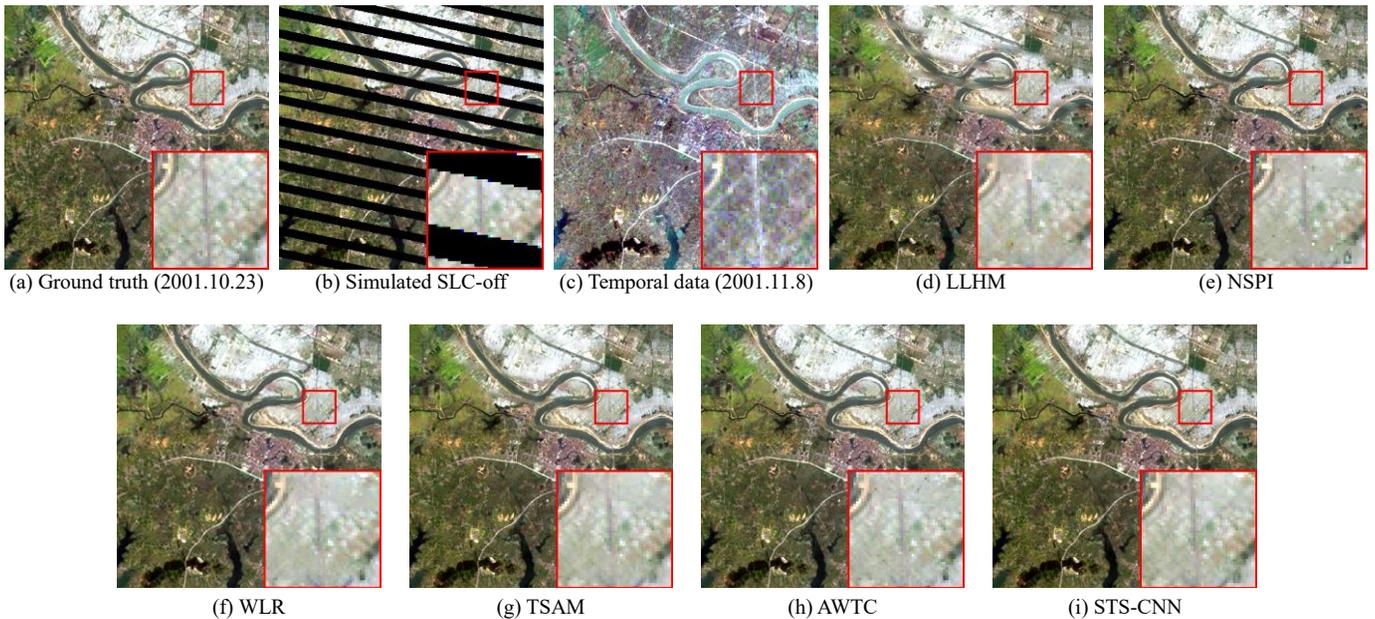

Fig. 11. Simulated ETM+ SLC-off recovery results with Landsat TM data.

TABLE II. QUANTITATIVE EVALUATION RESULTS OF THE SIMULATED SLC-OFF PROBLEM IN LANDSAT TM DATA.

|  | mPSNR (↑) | mSSIM (↑) | CC (↑) | SAM (↓) |
| --- | --- | --- | --- | --- |
| LLHM | 34.852 ± 0.024 | 0.910 ± 0.001 | 0.940 ± 0.001 | 1.842 ± 0.012 |
| NSPI | 36.533 ± 0.025 | 0.927 ± 0.002 | 0.957 ± 0.002 | 1.234 ± 0.008 |
| WLR | 37.616 ± 0.032 | 0.939 ± 0.001 | 0.966 ± 0.001 | 1.126 ± 0.010 |
| TSAM | 37.953 ± 0.023 | 0.944 ± 0.002 | 0.969 ± 0.001 | 1.107 ± 0.008 |
| AWTC | 37.984 ± 0.015 | 0.947 ± 0.001 | 0.970 ± 0.001 | 1.114 ± 0.006 |
| STS-CNN | **38.531 ± 0.017** | **0.950 ± 0.001** | **0.973 ± 0.001** | **1.013 ± 0.009** |

From the results in Fig. 10, LF shows obvious stripe noise along the dead lines, on account that the relationship between band 6 and band 7 relies on many complex factors, and is not a simple linear regression correlation. For the HMLLSF and WCLF methods, although the histogram matching or pre-classification based linear regression strategy can complete the dead pixels, some stripe noise still exists, such as the enlarged regions in Fig. 10(d)-(f). This is because the degraded image contains various object classes, within which also exist internal differences rather than homogeneous property in different regions. For AWTC method, although the weights in the different unfolding are adaptively determined, the weights of the different singular values are not taken into account. Meanwhile, for the proposed STS-CNN method, the dead lines are well recovered, and the local textures and overall tone are also well preserved, without generating obvious stripe noise, which can be clearly observed from the enlarged regions in Fig. 10(g). Furthermore, in terms of the three evaluation indexes in Table I, the proposed method also obtains better results than LF, HMLLSF, WCLF and AWTC.

**2) Simulated ETM+ SLC-Off in Landsat TM Data**

For the Landsat ETM+ SLC-off problem, we simulated this in TM data, as shown in Fig. 11(a) and (b), and employed a temporal image, as shown in Fig. 11(c). We show the simulated recovery results for the TM image through the six methods— LLHM [30], NSPI [31], WLR [32], TSAM [36], AWTC [4], and the proposed STS-CNN—in Fig. 11. To show the recovery results more clearly, enlarged parts of the reconstructing results are supplemented in Fig. 11, respectively. Furthermore, the quantitative evaluations with mSSIM, mPSNR, CC, and SAM are listed in Table II. As shown in Fig. 11(d)-(h), the comparative methods all result in discontinuous detail feature, to some degree. This is because different temporal data exist a highly complex non-linear relation, while the contrastive methods above didn't well fit this situation for missing data reconstruction. In comparison, the proposed STS-CNN model (Fig. 11(i)) performs better in reducing the spectral distortion, and shows a superior performance over the state-of-the-art methods in the quantitative assessment in Table II. This also verified the powerful non-linear expression ability of deep learning in the proposed method.

**3) Simulated Cloud Removal of Landsat Images**

Similar to the simulated experiment of ETM+ SLC-off, we also simulated the thick cloud removal task in TM data with multi-temporal data, as shown in Fig. 12(a), (b), and (c). The simulated recovery results for the TM image are shown in Fig. 12(d)-(i) for the six methods: LLHM [30], mNSPI [46], WLR [32], STWR [35], AWTC [4], and the proposed STS-CNN method, respectively. The quantitative evaluations are shown in Table III. Clearly, in Fig. 12(d)-(h), the results of LLHM, mNSPI, WLR, STWR and AWTC also show texture discontinuity or spectral distortion in some degree, because the relationship between different temporal data is not a simple linear correlation, but a highly complex non-linear correlation. In contrast, the proposed method performs well in reducing spectral distortion, and shows a nice performance in the quantitative assessment in Table III.

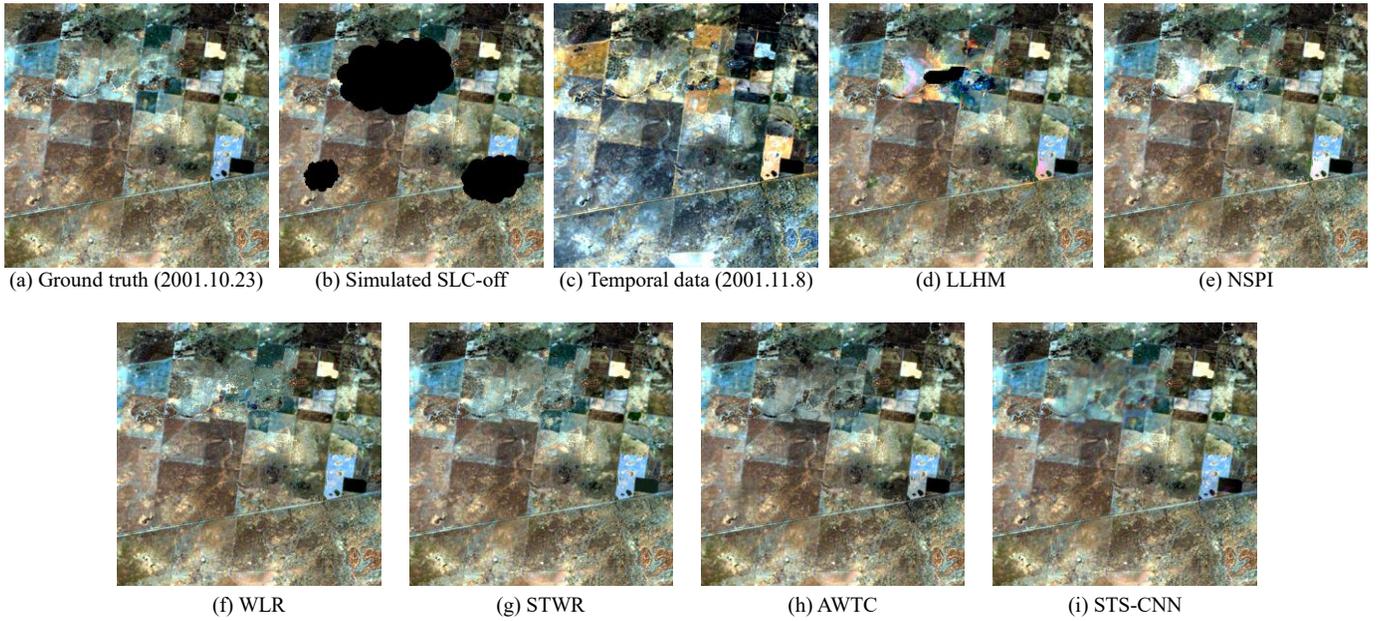

Fig. 12. Simulated recovery results of cloud removal in Landsat TM data.

TABLE III. QUANTITATIVE EVALUATION RESULTS OF THE SIMULATED CLOUD REMOVAL IN LANDSAT TM DATA.

|  | mPSNR (↑) | mSSIM (↑) | CC (↑) | SAM (↓) |
| --- | --- | --- | --- | --- |
| LLHM | 21.357 ± 0.022 | 0.883 ± 0.001 | 0.913 ± 0.002 | 2.230 ± 0.014 |
| mNSPI | 24.079 ± 0.016 | 0.895 ± 0.002 | 0.943 ± 0.002 | 1.324 ± 0.017 |
| WLR | 24.812 ± 0.018 | 0.897 ± 0.001 | 0.959 ± 0.003 | 1.281 ± 0.025 |
| STWR | **27.014 ± 0.015** | 0.910 ± 0.002 | 0.963 ± 0.002 | 0.775 ± 0.018 |
| AWTC | 25.973 ± 0.008 | 0.899 ± 0.001 | 0.960 ± 0.001 | 1.293 ± 0.009 |
| STS-CNN | 26.805 ± 0.014 | **0.912 ± 0.001** | **0.965 ± 0.002** | **0.766 ± 0.016** |

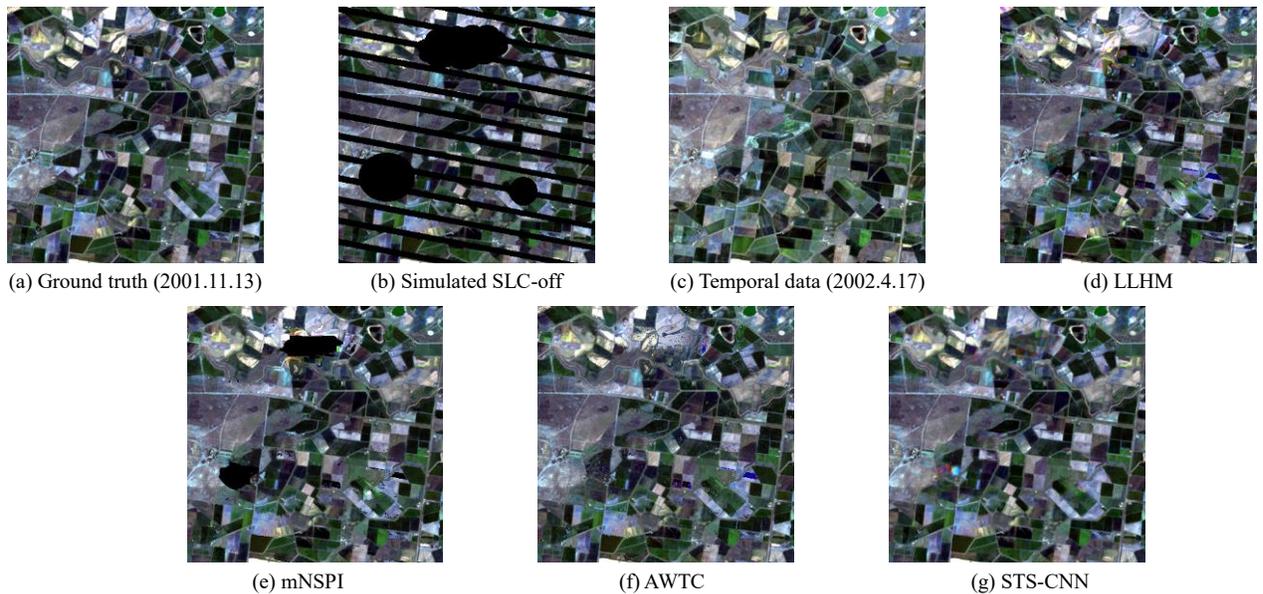

Fig. 13. Recovery results for the simulated Landsat TM data with both ETM+ SLC-off and thick cloud.

TABLE IV. QUANTITATIVE EVALUATION RESULTS FOR THE LANDSAT TM DATA WITH BOTH ETM+ SLC-OFF AND THICK CLOUD.

|  | mPSNR (↑) | mSSIM (↑) | CC (↑) | SAM (↓) |
| --- | --- | --- | --- | --- |
| LLHM | 18.447 ± 0.023 | 0.749 ± 0.002 | 0.882 ± 0.003 | 4.996 ± 0.028 |
| mNSPI | 19.545 ± 0.018 | 0.786 ± 0.001 | 0.894 ± 0.002 | 4.370 ± 0.022 |
| WLR | 23.920 ± 0.022 | 0.838 ± 0.001 | 0.932 ± 0.001 | 3.141 ± 0.024 |
| STS-CNN | **24.063 ± 0.016** | **0.853 ± 0.001** | **0.946 ± 0.002** | **2.990 ± 0.025** |

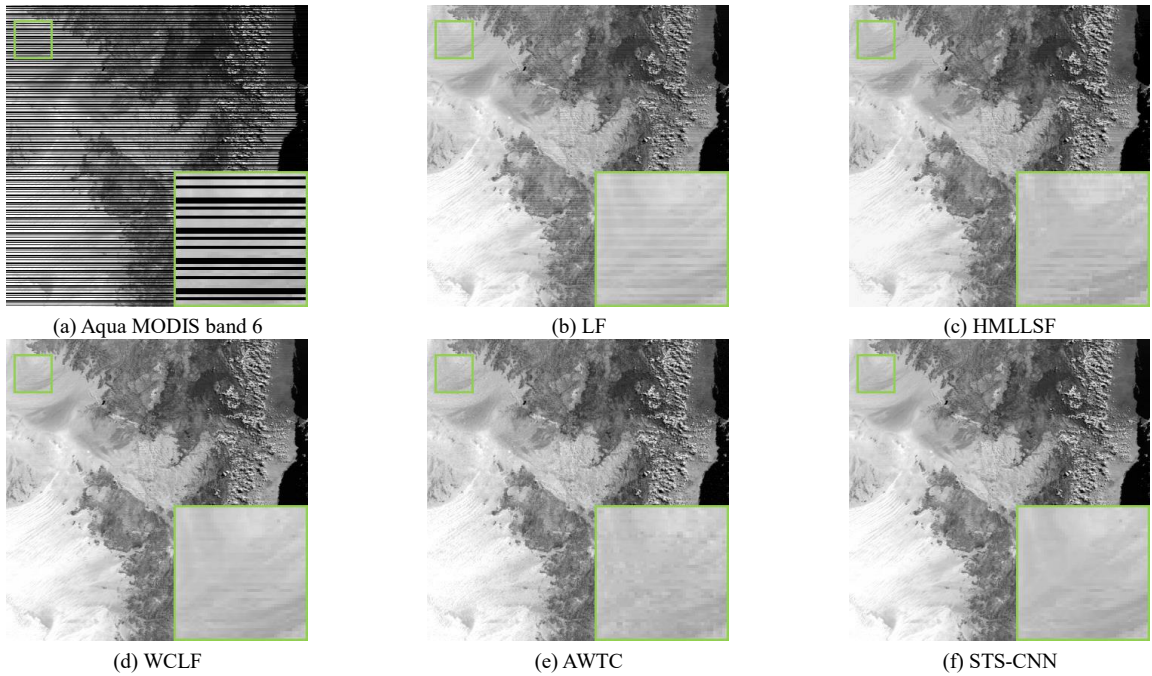

Fig. 14. Real recovery results for Terra MODIS band 6.

**4) Simulated TM Data with Both Cloud and SLC-Off**

Considering that SLC-off data may also contain thick cloud, a simulated experiment with both SLC-off and cloud cover was undertaken to verify the effectiveness of the proposed method. Fig. 13(d)–(g) shows the reconstruction outputs of LLHM, NSPI, WLR, and the proposed STS-CNN model, respectively. The quantitative evaluations with mSSIM, mPSNR, CC, and SAM are listed in Table IV.

Clearly, for the reconstruction of remote sensing data with both SLC-off and large missing areas, LLHM and mNSPI cannot completely recover the cloud-covered regions, and the result of WLR also shows texture discontinuity. In contrast, the proposed STS-CNN model performs better in reducing spectral distortion, and shows a better performance over the state-of-the-art methods in the quantitative assessment in Table IV.

### C. Real-Data Experiments

**1) Dead Lines in Aqua MODIS Band 6**

The results of the real-data experiment for reconstructing dead pixels in Aqua MODIS band 6 are shown in Fig. 14(b)–(f), including the outputs of LF, HMLLSF, WCLF, AWTC, and STS-CNN, respectively. From the overall visual perspective, all these methods can achieve reasonable outcomes with inconspicuous dissimilarities. However, some stripe noise is still found in the results of the comparative methods, as shown in the enlarged regions of Fig. 14(b)–(e).

For the HMLLSF and WCLF methods, although the histogram matching or pre-classification based linear regression strategy can complete the dead pixels, some stripe noise still exists, such as the enlarged regions in Fig. 14(b)-(e). This is because the degraded image contains various object classes, within which also exist internal differences rather than homogeneous property in different regions. For AWTC method, it also produced some artifacts in enlarged regions because of the complex relations between different bands. In contrast, the proposed STS-CNN model (Fig. 14(e)) can effectively recover the dead lines and simultaneously reduce the artifact detail, such as stripe noise, as shown in the marked regions of Fig. 14(f).

**2) SLC-Off ETM+ Images**

The results of the real-data experiment for reconstructing Landsat ETM+ SLC-off data are shown in Fig. 15, where Fig. 15(a) and (b) are the two temporal ETM+ SLC-off images observed on October 23, 2011, and November 8, 2011, respectively. Fig. 15(c)–(h) show the outputs of LLHM, NSPI, WLR, TSAM, AWTC, and the proposed STS-CNN method, respectively. Because the gaps cannot be completely covered by the single auxiliary SLC-off image, there are still invalid pixels remaining, so the land parameter retrieval model (LPRM) [42] algorithm was employed after the processing of LLHM, NSPI, and WLR. However, the proposed STS-CNN model does not require LPRM to complete the residual gaps, as a result of the end-to-end strategy. From the overall visual perspective, all the methods can fill the gaps. However, for the five contrastive methods, some stripe noise can still be observed. In comparison, the proposed method can both recover the gaps and shows the least stripe noise, as shown in Fig. 15(h). Furthermore, for the five other contrast methods, some detail texture is inconsistent or discontinuous in the

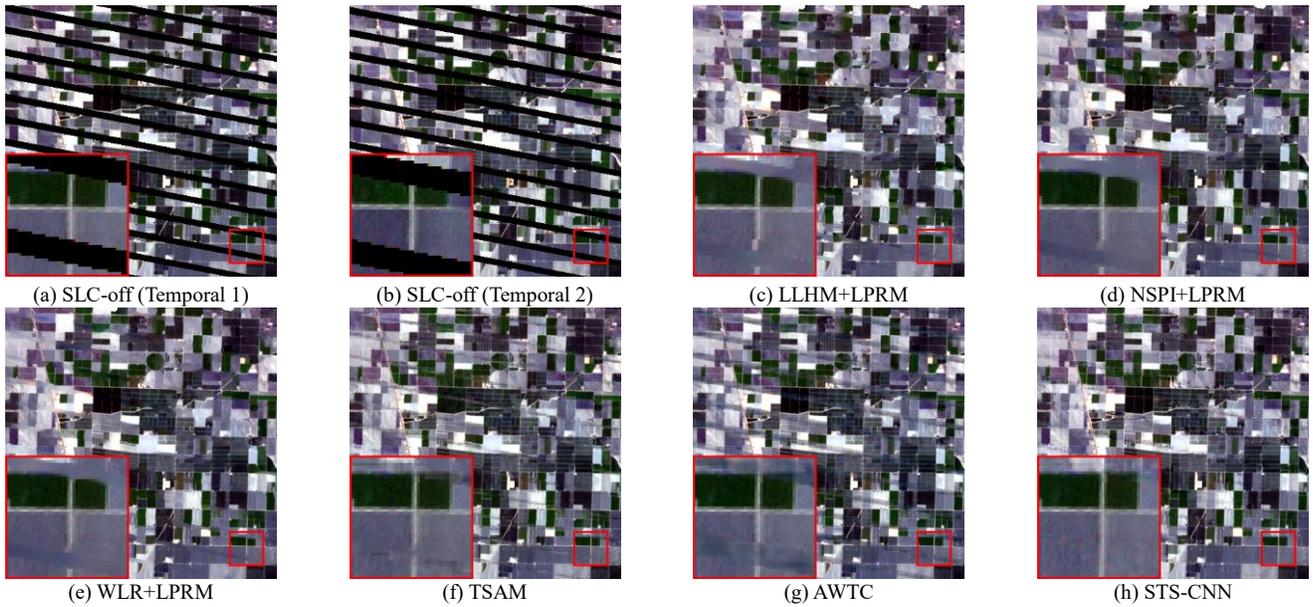

Fig. 15. SLC-off reconstruction results for the real ETM+ SLC-off image.

(a) SLC-off (Temporal 1)  (b) SLC-off (Temporal 2)  (c) LLHM+LPRM  (d) NSPI+LPRM
(e) WLR+LPRM  (f) TSAM  (g) AWTC  (h) STS-CNN

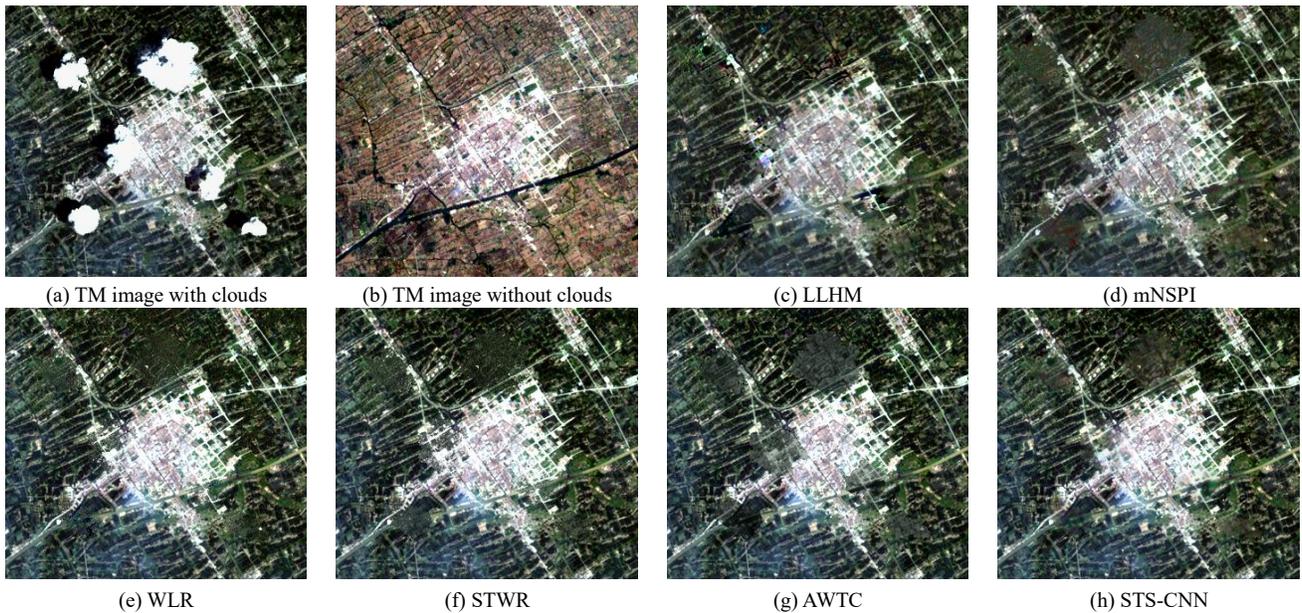

(a) TM image with clouds  (b) TM image without clouds  (c) LLHM  (d) mNSPI
(e) WLR  (f) STWR  (g) AWTC  (h) STS-CNN

Fig. 16. Real-data recovery results for cloud removal in Landsat TM data.

reconstruction regions of the dead lines, which can be clearly observed from the enlarged region. Meanwhile, STS-CNN can simultaneously preserve the detail texture and acquire a much more consistent and continuous reconstruction result for the dead pixels.

**3) Cloud Removal of TM Images**

The results of the real-data experiment for recovering a TM image with thick cloud are shown in Fig. 16(a)–(h), where Fig. 16(a) and (b) are the two temporal TM images which contained thick cloud. Fig. 16(c)–(h) show the reconstruction outputs of LLHM, NSPI, WLR, STWR, AWTC and the proposed STS-CNN method, respectively.

For LLHM, mNSPI, and AWTC, it can be clearly observed that areas within the largest cloud contain some spectral distortion. Besides, for reconstructing remote sensing data of large missing areas, the results of mNSPI, WLR, and AWTC also show texture discontinuity, because the relationship between different temporal data is not a simple linear correlation, but a complex non-linear correlation. In addition, for the WLR and STWR method, the reconstruction texture details of cloud areas are not inconsistent with no cloud areas around, which cannot fit the non-linear relation between different temporal data. In contrast, the proposed STS-CNN method performs better in reducing spectral distortion, and shows a nice performance in the quantitative assessment in Table III. For the proposed STS-CNN model, the texture details are better reconstructed than for WLR and STWR, and the spectral distortion is less than for LLHM, mNSPI and AWTC.

## D. Further Discussion

### 1) Analysis of the Proposed Network Components

To verify the validity of the proposed model structure, three pairs of comparison experiments with two simulated images were carried out, as shown in Fig. 10 (simulated experiment for dead lines in Terra MODIS band 6) and Fig. 11 (simulated experiment for ETM+ SLC-off), respectively. Fig. 17 shows the PSNR or mPSNR values obtained with/without: 1) the multi-scale feature extraction block; 2) dilated convolution; and 3) boosting of the spatial and temporal/spectral information under the same setting environment. The iterations for all six experiments were set to 500 000, and training models were extracted every thousand iterations for testing.

For the multi-scale feature extraction block, we can observe that it can promote the accuracy of the reconstruction by about 1/0.5 dB, as shown in Fig. 17(a) and (b), respectively, indicating that extracting more features with multi-context information is beneficial for restoring missing regions. For the dilated convolution, Fig. 17(c) and (d) also confirm its effectiveness, with a promotion of 0.2/0.4 dB, respectively. With regard to the boosting of the spatial and temporal/spectral information, Fig 17(e) and (f) clearly demonstrate its effects on the spatial and temporal/spectral information transfer in the proposed STS-CNN model.

### 2) Effects of Image Registration Errors

For pairs of temporal data, it should be stressed that registration errors cannot be ignored, and they can affect the reconstruction results, to some extent. Therefore, we set registration errors of 0–5 pixels in series for the simulated ETM+ SLC-off experiment with LLHM, NSPI, WLR, and the proposed STS-CNN method. Fig. 18(a)–(d) shows the reconstruction results of the comparing algorithms with registration errors of 2 pixels, respectively. In addition, four broken line graphs of the four methods are shown in Fig. 19(a)–(d), demonstrating the tendency of mSSIM, mPSNR, CC, and SAM with the registration errors, respectively. Clearly, the proposed method still obtains better recovery results when compared with LLHM, NSPI, and WLR, as shown in Fig. 18. As the image registration errors increase, the degradation rate of the proposed method is the lowest, compared with LLHM, NSPI, and WLR. One possible reason for this may be that these linear models are heavily dependent on the corresponding relationship of neighborhood pixels, whose reconstruction accuracy is seriously restricted by image registration errors. In contrast, the recovery method based on a deep CNN can effectively take advantage of its powerful non-linear expression ability, and can enlarge the size of the contextual information, which can help to resist or reduce the negative impact of image registration errors, as can be observed in Fig. 19.

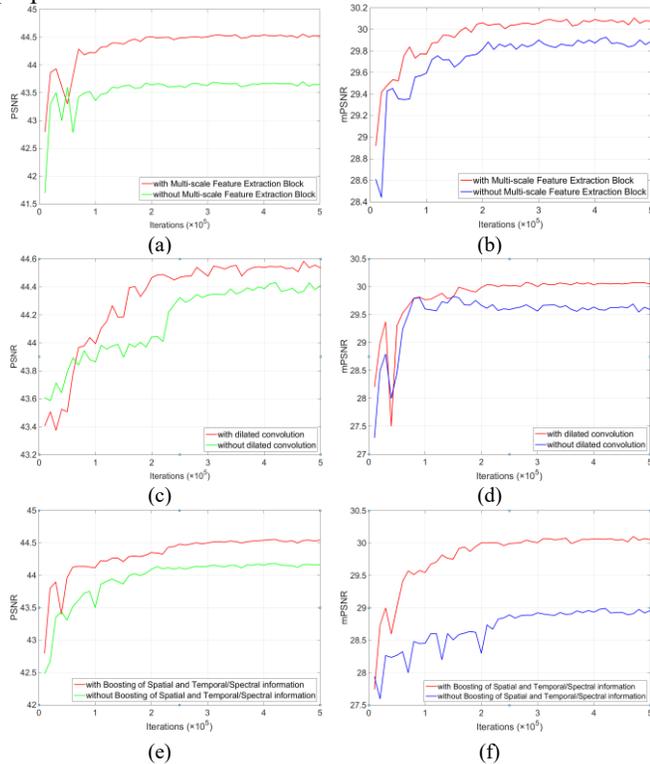

Fig. 17. Analysis of the effectiveness of the proposed network structure components. (a) With/without the multi-scale feature extraction block in Fig. 10. (b) With/without the multi-scale feature extraction block in Fig. 11. (c) With/without the dilated convolution in Fig. 10. (d) With/without the dilated convolution in Fig. 11. (e) With/without the boosting of the spatial and temporal/spectral information in Fig. 10. (f) With/without the boosting of the spatial and temporal/spectral information in Fig. 11.

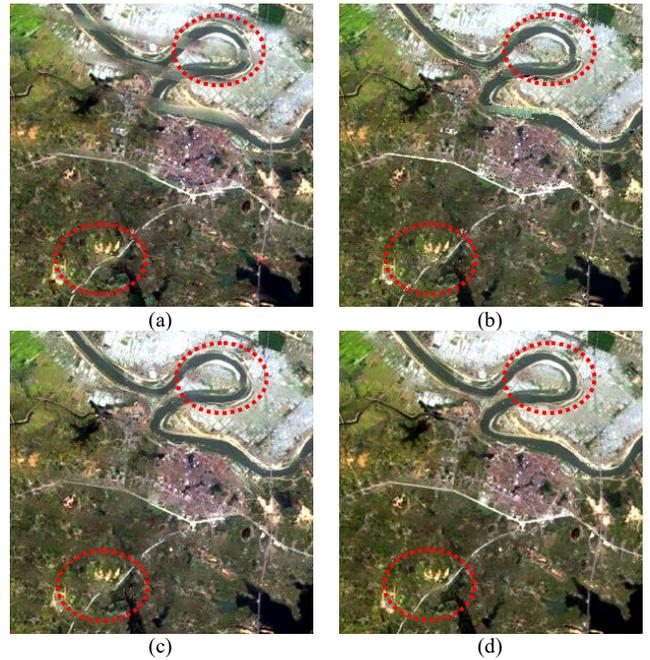

Fig. 18. Reconstruction example with a 2-pixel image registration errors. (a) LLHM. (b) NSPI. (c) WLR. (d) STS-CNN.

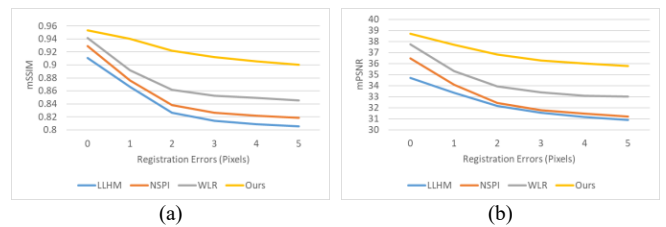

(a)          (b)

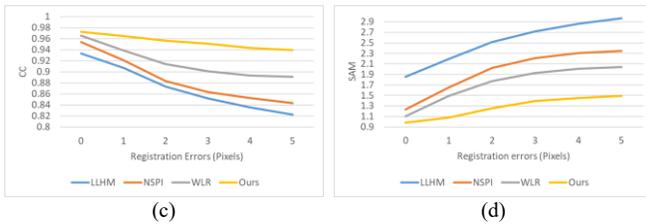

(c)                       (d)

Fig. 19. Analysis of the effect of image registration error on the reconstruction. (a) mSSIM. (b) mPSNR (dB). (c) CC. (d) SAM.

## V. CONCLUSIONS

In this paper, we have presented a novel method for the reconstruction of remote sensing imagery with missing data, through a unified spatial-temporal-spectral framework based on a deep CNN. Differing from most of the inpainting methods, the proposed STS-CNN model can recover different types of missing information, including the dead lines in Aqua MODIS band 6, the Landsat SLC-off problem, and thick cloud removal. It should be noted that the proposed model uses multi-source data (spatial, spectral and temporal information) as the input of the unified framework. Furthermore, to promote the reconstruction precision, some specific structures are employed in the network to enhance the performance. Compared with other traditional reconstruction methods, the results show that the proposed method shows a significant improvement in terms of reconstruction accuracy and visual perception, in both simulated and real-data experiments.

Although the proposed method performs well for reconstructing the dead lines in Aqua MODIS band 6, the ETM+ SLC-off problem, and thick cloud removal, it still has some unavoidable limitations. When removing thick cloud through the use of temporal information, it results in some spectral distortion and blurring. Another possible strategy which will be explored in our future research like adding *a priori* constraint (such as neighborhood similar pixels interpolation [47], locality adaptive discriminant analysis [48], embedding structured contour and location [49], and context transfer [50] etc.) to reduce the spectral distortion and improve the texture details.

*Geosci. Remote Sens.*, vol. 52, no. 1, pp. 175–187, Jan. 2014.